\documentclass{article}
\usepackage{ijcai24}

\usepackage{times}
\usepackage{soul}
\usepackage{url}
\usepackage[hidelinks]{hyperref}
\usepackage[utf8]{inputenc}
\usepackage[small]{caption}
\usepackage{graphicx}
\usepackage{amsmath}
\usepackage{amsthm}
\usepackage{amssymb}
\usepackage{booktabs}
\usepackage{algorithm}
\usepackage{algorithmic}
\usepackage{subfigure}
\usepackage[switch]{lineno}

\newtheorem{theorem}{Theorem}
\newtheorem{assumption}{Assumption}
\newtheorem{definition}{Definition}

\title{Estimating Conditional Average Treatment Effects via  \\
           Sufficient Representation Learning}

\author{
Pengfei Shi$^{1,3}$
\and
Wei Zhong$^1${\footnote{
Corresponding author}} \and
Xinyu Zhang$^2$\and
Ningtao Wang$^{3}$\footnotemark[1] \and
Xing Fu$^3$ \and 
Weiqiang Wang$^3$\and
Yin Jin$^3$ \\
\affiliations
$^1$Xiamen University, Xiamen, China\\
$^2$Academy of Mathematics and Systems Science, Chinese Academy of Sciences, Beijing, China\\
$^3$Ant Group, Hangzhou, China\\
\emails
\{lasou.spf, ningtao.nt, zicai.fx, weiqiang.wwq, jinyin.jin\}@antgroup.com,
wzhong@xmu.edu.cn,
xinyu@amss.ac.cn
}

\begin{document}

\maketitle

\begin{abstract}
    Estimating the conditional average treatment effects (CATE) is very important in causal inference and has a wide range of applications across many fields. In the estimation process of CATE, the unconfoundedness assumption is typically required to ensure the identifiability of the regression problems. When estimating CATE using high-dimensional data, there have been many variable selection methods and neural network approaches based on representation learning, while these methods do not provide a way to verify whether the subset of variables after dimensionality reduction or the learned representations still satisfy the unconfoundedness assumption during the estimation process, which can lead to ineffective estimates of the treatment effects. Additionally, these methods typically use data from only the treatment or control group when estimating the regression functions for each group. This paper proposes a novel neural network approach named \textbf{CrossNet} to learn a sufficient representation for the features, based on which we then estimate the CATE, where \textbf{cross} indicates that in estimating the regression functions, we used data from their own group as well as cross-utilized data from another group. Numerical simulations and empirical results demonstrate that our method outperforms the competitive approaches.
\end{abstract}

\section{Introduction}
\label{intro}
Estimating the CATE, also known as individual treatment effects (ITE) or heterogeneous treatment effects (HTE), is of great importance in causal inference and is widely applied in numerous fields \cite{lalonde1986evaluating,athey2016recursive,mooij2016distinguishing,yao2021survey}. In the medical field, the estimation of CATE can be applied to precision medicine, achieving personalized matching of treatment plans. In the socio-economic domain, the setting of policies and pre-assessment of their effects also rely on the estimation of treatment effects. In the field of advertising recommendations, making accurate recommendations of different ads to different groups can effectively save costs and enhance the effectiveness of the advertising. All these depend on the accurate estimation of CATE.

The estimation of the CATE is typically divided into randomized experimental studies \cite{whitmore2005resource,rosenbaum2007interference} and observational data studies, with the distinction being whether the allocation of interventions is completely random. Due to the constraints of real-world settings and specific problems, we are often unable to conduct randomized experimental studies. For example, when analyzing the impact of smoking on lung cancer, we cannot ethically require some participants to smoke over an extended period. On the other hand, randomized experiments also come with high costs. Therefore, most researches are based on observational data, and this article is no exception. Most of the literature on research based on observational data operates under the assumption of unconfoundedness:
\begin{assumption}[Unconfoundedness]
    Given the observed variables, the treatment variable is independent with the potential outcomes: $Y(1),Y(0)\perp T \mid \mathbf{x}$.
\end{assumption}

\begin{figure*}[t]
    \centering 
    \subfigure[General network structure]{%
        \centerline{\includegraphics[width=1.1\columnwidth]{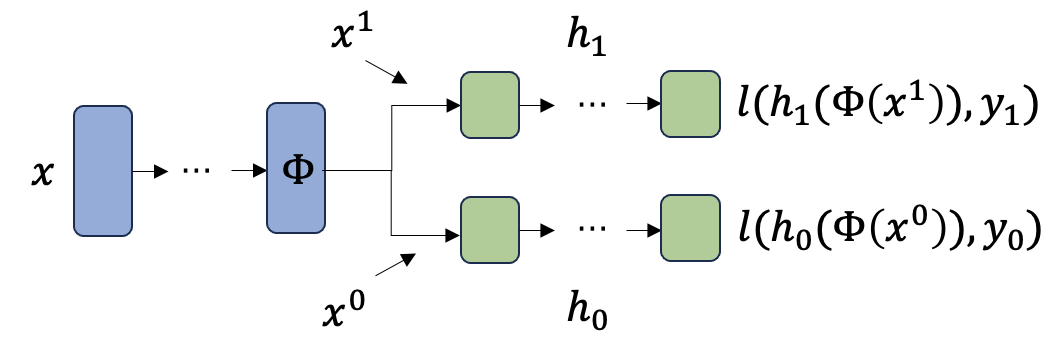}}%
    }
    
    \vspace{-2mm}
    
    \subfigure[Our network structure]{%
        \centerline{\includegraphics[width=1.7\columnwidth]{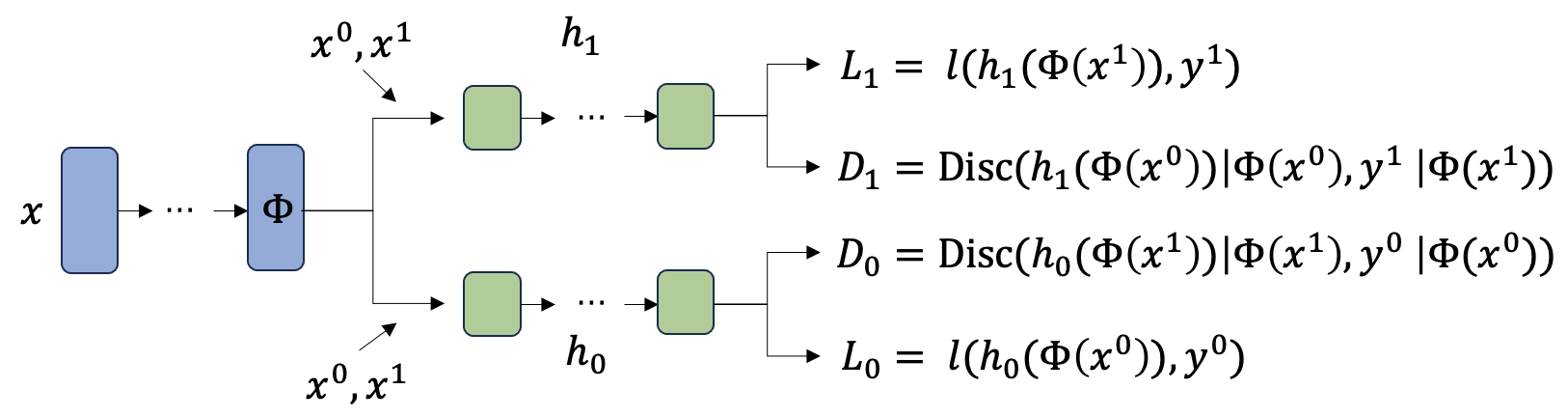}}%
    }
    \caption{$\Phi$ is a learned representation. $h_1$ and $h_0$ are two hypothesis or predictive functions for $y$ based on $\Phi$ in treated and control group respectively. $l(\cdot)$ is a loss function, where we use mean squared error loss. $\mbox{Disc}(\cdot)$ means the discrepancy. $x^1,y^1$ and $x^0,y^0$ are the covariates and responses corresponding to treated group sample and control group sample respectively.}
    \label{structure}
\end{figure*}
\noindent The meaning of this assumption is that the features we can collect contain all the information that could simultaneously affect both the treatment variable and the outcome variable, thereby creating a situation similar to a randomized experiment, where the allocation of interventions is completely random given these features. On the other hand, since we can only observe one of the two potential outcomes, we can not directly identify the CATE from the observational data unless the assumption of unconfoundedness holds. As we collect more data and the dimensionality of features increases, it becomes easier for the unconfoundedness assumption to be satisfied \cite{rolling2014model}. However, when estimating CATE using high-dimensional data, many studies perform variable selection \cite{vanderweele2019principles} or representation learning \cite{shalit2017estimating,shi2019adapting,hassanpour2019counterfactual,hassanpour2019learning,zhang2021treatment,chauhan2023adversarial} on the original data, and the resulting estimates of the treatment effect are based on the selected subset of variables or learned representations as shown in Figure \ref{structure} (a). These methods do not verify whether the representations or the selected subset of variables still meet the unconfoundedness assumption, nor do they make additional assumptions, that is, we are unable to confirm whether $Y(1),Y(0)\perp T \mid \Phi(\mathbf{x})$ holds or not. If the unconfoundedness assumption no longer holds, the estimates of treatment effects based on $\Phi(\mathbf{x})$ are also no longer valid. Therefore, we propose a new type of neural network structure that, by imposing specific regularization terms, to ensure the learned representations still satisfy the unconfoundedness assumption, thus making the final treatment effect estimates accurate and valid.

On the other hand, most of the neural network approaches are conducted under the T-learner framework, which involves estimating a separate regression function for the treated group and the control group respectively, and then obtaining a plug-in estimator by taking the difference between the two regression functions. However, as shown in Figure \ref{structure} (a), in training the regression functions for the treatment and control groups, they typically use data from their respective groups, that is, the treated group's hypothesis function is estimated using data from the treated group, and the control group's hypothesis function is estimated using data from the control group, which can lead to an underutilization of data. Actually, the goal is to estimate the treatment effect, not to achieve highly accurate estimates of the two hypothesis functions. Due to selection bias, having precise estimates for the two hypothesis functions does not always lead to an accurate estimate of the treatment effect. If we train the models using data from each group independently, we will lose the cross-information between groups. This results in the regression function of the treatment group not being well-suited for the data of the control group, and vice versa. \cite{kunzel2019metalearners} proposed X-leaner, where they use data from another group to evaluate the hypothesis functions and then take average of the treatment effect under treated and control groups. However, they do not apply the data from another group into the training process in fact. The method we propose makes use of the full sample information when training the hypothesis functions. Through the design of the objective function, the treatment and control group data play different roles, aiming to achieve different optimization effects. The main contributions in this paper are as follows:
\begin{itemize}
    \item We define a sufficient representation in causal inference, which not only has strong predictive power for potential outcomes, but also needs to satisfy the unconfoundedness assumption. We propose a novel objective function that enables us to learn the sufficient representations.
    \item During the training of the hypothesis functions, we use data from their own group as well as cross-utilize data from another group, allowing the two learned hypothesis functions to have good predictive capability for the counterfactual outcomes, thereby obtaining a more accurate estimate of the treatment effect.
    \item The regularization terms in our method take into account the information of the conditional distributions of counterfactual outcomes, which reduces the selection bias effectively and is more in line with the intuition of causal inference. 
\end{itemize}
The rest of this paper is organized as follows. In Section 2, we discuss related works. Our proposed approach is represented in Section 3. In Section 4, we conduct Monte Carlo simulation and experiments. Finally, we give a conclusion in Section 5.

\section{Related Work}
\cite{kunzel2019metalearners} proposed the concept of metalearners for estimating CATE, which decomposes the problem of estimating CATE into multiple sub-problems, each of which can be addressed using any supervised or regression machine learning method. \cite{kunzel2019metalearners} was the first to combine metalearners with CATE, categorizing the previous methods into two types. The first type is S-Learner, where “S” denotes single. We can model the treatment variable together with other covariates using a single model, where only one estimator for $\mu(x,t)=\mathbb{E}[Y| X=x,T=t]$ is obtained. Then, the estimate for CATE can be written as $\hat{\tau}_S(x)=\hat{\mu}(x,1)-\hat{\mu}(x,0)$. Under this framework, \cite{hill2011bayesian}, \cite{green2012modeling} conducted research with Bayesian Additive Regression Trees (BARTs) as the baseline model. \cite{athey2016recursive} explored this further using tree-based models as the benchmark. The biggest shortcoming of the S-learner is that it does not take into account the interaction between covariates and the treatment variable, that is, it does not allow for the hypothesis functions of the treatment and control groups to have different forms.

The second type is T-learner, where “T” denotes two. We can model the treatment group and control group separately, obtaining estimates for two regression functions $\mu_1(x)=\mathbb{E}[Y(1)| X=x]$ and $\mu_0(x)=\mathbb{E}[Y(0)| X=x]$. By further subtracting them, we can obtain the estimate $\hat{\tau}_T(x)=\hat{\mu}_1(x)-\hat{\mu}_0(x)$ for CATE. Under this framework, \cite{hansen2009attributing} and \cite{wager2018estimation} proposed causal forest, where they used random forest as the benchmark. In recent years, there have also been many methods based on representation learning and neural networks, which are also a type of T-learner. \cite{shalit2017estimating} proposed the CFRNet, a neural network architecture that first learns a representation, and then by imposing a penalty on the integral probability metric measure of the representation between the treated and control groups, the model mitigates selection bias through a balanced representation learning. This work derived a bound for the expected error of the estimator for CATE given a fixed representation, where the hypothesis functions are defined over this representation. The key assumption is that this representation is a one-to-one function, so that the unconfoundedness assumption holds given this representation. This is a strong condition, as not all network architectures are invertible during the process of representation learning. \cite{shi2019adapting} proposed Dragonnet based on the sufficiency of the propensity score for treatment effect estimation. The neural network architecture learns a shared representation, which is then used to model the two hypothesis functions and the propensity scores, ensuring that this representation simultaneously contains information that can predict both the outcomes and the treatment variable. \cite{hassanpour2019learning} proposed a context-aware importance sampling re-weighing scheme, built on top of a representation learning module. In \cite{shalit2017estimating}, the weight of the loss is simply based on the sample size, which is $w_i=1/\mbox{Pr}(t_i=1)$. \cite{hassanpour2019counterfactual} further eliminates selection bias by estimating the nominal distribution of the representation, followed by importance sampling re-weighting. They use $1+\mbox{Pr}(\phi_i\mid 1-t_i)/\mbox{Pr}(\phi_i\mid t_i)$ as the loss weight. Another category of neural network approaches estimates the CATE by learning a disentangled representation, or disentangled latent factors. \cite{zhang2021treatment} proposed a data-driven algorithm TEDVAE to infer latent factors and disentangled them into three disjoint sets, which are to predict treatment, outcome, or both. \cite{hassanpour2019learning} also divided the information contained in all the covariates into three non-overlapping parts, characterizing them by learning three disentangled representations, and then modeled the treatment and outcome based on these representations. \cite{curth2021nonparametric}, \cite{chauhan2023adversarial} further divided the features into five parts, corresponding to treatment $t$, treated group outcomes $\mu_1$, control group outcomes $\mu_0$, both $\mu_1$ and $\mu_0$, and all of treatment and outcomes. 

\cite{kunzel2019metalearners} proposed the third metaalgorithm: X-leaner, which first used the T-learner approach to estimate two hypothesis functions. Then, by cross-utilizing the data, they substituted data from the other group into the hypothesis function to predict the counterfactual outcomes. By comparing these predictions with the factual outcomes, we can obtained the estimates of the individual treatment effects for all samples, which is $\Tilde{D}_i^1=Y_i^1-\hat{\mu}_0(X_i^1)$, $\Tilde{D}_i^0=\hat{\mu}_1(X_i^0)-Y_i^0$, where $\Tilde{D}^1$ is the predicted ITE for treated group, and $\Tilde{D}_i^0$ is the predicted ITE for control group. Subsequently, by regressing these individual effects against the individuals' characteristics, we can obtain the estimates of the treatment effects under treated group (ITT) $\hat{\tau}_1(x)$ and control group (ITC) $\hat{\tau}_0(x)$. Finally, they take a weighted average of these two estimates, resulting in the estimate for CATE $\hat{\tau}_X(x)=g(x)\cdot \hat{\tau}_1(x)+(1-g(x))\cdot \hat{\tau}_0(x)$. This method also cross-utilizes the data from different groups, but does not participate in the training process of their respective hypothesis functions, that is, the estimation of $\mu_1(x)$ and $\mu_0(x)$. In fact, when the unconfoundedness assumption is satisfied, $\hat{\tau}_0(x)$ and $\hat{\tau}_1(x)$ should be equal, which we will mention in the next section. \cite{kunzel2019metalearners} averages $\hat{\tau}_0(x)$ and $\hat{\tau}_1(x)$ using weights, while our approach encourages equality between them through a regularization term. This allows the data from different groups to participate in the training process of each other's hypothesis functions, since the estimation for $\hat{\tau}_0(x)$ and $\hat{\tau}_1(x)$ requires data from both treated and control groups. Our method fully utilizes the information from all the data and ensures that the unconfoundedness assumption is met given the learned representation.


\cite{curth2021nonparametric} summarizes lots of nonparametric statistical methods and machine learning approaches, from theoretical research to experiment, and supplements the field of CATE with meta-learners. The author collectively refers to the aforementioned S-learner and T-learner as one-step methods. That is, they estimate the conditional expectation only once and then use a plug-in method to estimate the CATE. The author defines a two-step method, which first involves estimating the nuisance parameters $\eta=\left(\mu_0(x), \mu_1(x), \pi(x)\right)$ to obtain $\hat{\eta}$, then regressing the pseudo-response variable $\tilde{Y}_{\hat{eta}}$, which depends on the nuisance parameters, on $x$. Here we consider the pseudo-response variable that satisfies $\mathbb{E}_\mathbb{P}[\tilde{Y}_\eta\mid X=x]=\tau(x)$. Based on this idea, \cite{curth2021nonparametric} proposed the RA-learner, which has similarities to X-learner \cite{kunzel2019metalearners}. At the same time, the author introduces the PW-learner, based on the Inverse Probability Weighting method (IPW). Lastly, the two-step method also includes the DR-learner, proposed by \cite{kennedy2020towards}, which is an extension based on the idea of doubly robust and adaptive Inverse Probability of Treatment Weighting \cite{robins1995semiparametric}. Our approach falls under the one-step approach, where we only need to estimate $\mu_1$ and $\mu_0$ and then obtain the final estimate by taking the difference. Unlike previous methods, we apply specific regularization terms to ensure that the training process makes full use of the data and that the learned representation satisfies the unconfoundedness assumption.

\section{Method}
We conduct our research based on the Rubin-Neyman potential outcomes framework \cite{rubin2005causal}. Denote $\mu_1(x)=\mathbb{E}(Y(1)| X=x)$ and $\mu_0(x)=\mathbb{E}(Y(0)| X=x)$, where $Y(1)$ and $Y(0)$ are two potential outcomes corresponding to the individual under treatment or control, while we can only observe one of them. Let $t\in \{0,1\}$ be a binary treatment variable, then the response we can observe for the $i$-th unit is $Y_i(t_i)$. Our goal is to estimate CATE, which denoted as $\tau(x):=\mathbb{E}(Y(1)-Y(0)|x)=\mu_1(x)-\mu_0(x)$. $\tau(x)$ is the individualized treatment effect, which depends on the characteristics $x$. Since we can only observe one of the two potential outcomes for each unit, to ensure identifiability of this problem, we need to impose strong ignorability condition, which is Assumption 1, along with the overlapping condition $0<\mbox{Pr}(t=1\mid x)<1$ for all $x$. Strong ignorability condition is commonly employed in works that are based on observational data. As the dimensionality of collectable features increases, this condition tends to be more easily satisfied. Beyond this, we do not make any additional assumptions. 

\subsection{Crossnet}
Similar to most neural network approaches, we first learn a representation $\Phi(x)$ from the original features, and then, based on this representation, we train two separate hypothesis functions $h_1(\Phi(x))$ and $h_0(\Phi(x))$ for the treated and control groups, respectively to estimate $\mu_1(x)$ and $\mu_0(x)$. Note that the unconfoundedness assumption is imposed on the original covariates $x$ not the representation $\Phi(x)$, which means that we cannot directly learn two hypothesis functions given any arbitrary representation. If $\Phi(\cdot)$ is a invertible map \cite{shalit2017estimating}, we have that 
\begin{align}
    \tau(x)&=\mathbb{E}(Y(1)-Y(0)\mid X=x) \nonumber\\
    &=\mathbb{E}(Y(1)-Y(0)\mid X=\Phi^{-1}(\Phi(x))) \nonumber\\
    &=\mathbb{E}(Y(1)-Y(0)\mid \Phi(X)=\Phi(x)).
\end{align}
Thus, learning two predictive functions for $Y(1)$ and $Y(0)$ based on $\Phi(x)$ is reasonable and sufficient. However, when training the model in practice, it is not guaranteed that the learned representation is invertible, meaning a one-to-one mapping, or enforcing the condition of an invertible representation might significantly reduce the model's performance and effectiveness. Therefore, we cannot ensure that the predictive functions estimated based on the representation are reasonable and effective. Another viable approach is to impose the unconfoundedness assumption on the learned representation itself, i.e., condition $Y(1),Y(0)\perp T \mid \Phi(X)$. This ensures that the predictive functions trained based on the representation are also reasonable and effective. However, since the representation is a result of the network's learning process and possesses randomness, we cannot directly impose the condition on this stochastic representation. In this paper, we propose to learn a sufficient representation for $x$, which satisfies the unconfoundedness assumption, and is sufficient for explaining the outcomes. First we give the definition of sufficient representation in the estimation of CATE.
\begin{definition}[Sufficient representation]
\label{def1}
    Let $\Phi:\mathcal{X}\rightarrow\mathcal{R}$ be a representation function, where $\mathcal{R}$ is the representation space. If $\Phi(x)$ is a sufficient representation for $x$, $\Phi(x)$ should satisfy that
    \begin{itemize}
        \item [i)] $Y(1),Y(0)\perp T \mid \Phi(x)$,
        \item [ii)] $\mathbb{E}[Y(1)|X=x]\perp Y \mid \Phi(x)$ and $\mathbb{E}[Y(0)|X=x]\perp Y \mid \Phi(x)$.
    \end{itemize}
\end{definition}
\noindent In the above definition, i) means that the representation should satisfy the unconfoundedness assumption, so that the regression of $Y$ upon $\Phi(x)$ is reasonable for treatment effects. ii) means that the representation is sufficient for predicting the potential outcomes. Most of the aforementioned works only consider the second condition, that is the explanatory ability for the outcomes. Since $T$ is binary, we can rewrite i) as 
\begin{align}
    &~\mbox{Pr}(Y(1)|T=1,\Phi(x))= \mbox{Pr}(Y(1)|T=0,\Phi(x)),\nonumber\\
    &~\mbox{Pr}(Y(0)|T=0,\Phi(x))=\mbox{Pr}(Y(0)|T=1,\Phi(x)),
\end{align}
where $\mbox{Pr}(\cdot)$ is the probability density function. This condition means that the distribution of two potential outcomes conditional on $\Phi(x)$ is the same across different groups. Therefore, we can learn the representation $\Phi(\cdot)$ by encouraging the estimated conditional distribution for $Y(1)|\Phi(x)$, $Y(0)|\Phi(x)$ under treated group and control group to be consistent. Let $x^1,y^1$ and $x^0,y^0$ be the covariates and responses corresponding to treated group sample and control group sample respectively. As shown in Figure \ref{structure} (b), after obtaining the estimated hypothesis functions $h_1(\cdot)$ and $h_0(\cdot)$ for treated and control group, we can input another group of data into the hypothesis function to obtain the estimated counterfactual outcomes for all individuals in that group, that is $h_1(\Phi(x^0))$ and $h_0(\Phi(x^1))$. Then we can calculate the empirical conditional distribution $y_a^b|\Phi(x^a)$, where $a$ denotes the group that this individual actually belongs to, and $b$ is the group, for which we calculate the potential outcome. Specifically, 
\begin{align}
    \label{ITTITC}
    &y_0^0=y^0,~y_1^0=h_0(\Phi(x^1)),\nonumber\\
    &y_1^1=y^1,~y_0^1=h_1(\Phi(x^0)).
\end{align}
Then, we can calculate the discrepancy between the conditional distributions $y_0^b| \Phi(x^0)$ and $y_1^b| \Phi(x^1)$, for $b=0,1$, denoted as $\mbox{Disc}(p(y_0^b| \Phi(x^0)), p(y_1^b| \Phi(x^1)))$. Denote $l(\cdot)$ as a loss function, where we use mean squared error. Then, the optimization problem we propose is
\begin{align}
    \min_{\Phi,h_1,h_0}&l(h_1(\Phi(x^1)),y^1)+l(h_0(\Phi(x^0)),y^0) \nonumber\\
    &+\lambda\cdot \mbox{Disc}(p(y_0^0| \Phi(x^0)), p(y_1^0| \Phi(x^1))) \nonumber\\
    &+\lambda\cdot \mbox{Disc}(p(y_0^1| \Phi(x^0)), p(y_1^1| \Phi(x^1))),
\end{align}
where we can choose to use different discrepancy measure such as the integral probability metric (IPM), KL divergence, conditional MMD based on the nature of the data we encounter. In this paper, we use the statistic proposed by \cite{yu2021measuring}, which measures the divergence from $p_1(y| x)$ to $p_2(y| x)$, without estimating the distribution. Denote that $\mathcal{S}_+^n=\{A\in \mathbb{R}^{n\times n}\mid A=A^T,A\succcurlyeq 0\}$, then the Bregman matrix divergence \cite{kulis2009low} from the matrix $A$ to the matrix $B$ is defined as 
\begin{align}
    D_{\varphi}(A \| B)=\varphi(A)-\varphi(B)-\operatorname{tr}\left((\nabla \varphi(B))^T(A-B)\right),
\end{align}
where $\varphi(\cdot)$ is a strictly convex, differentiable function, that maps matrices to the extended real numbers. The divergence from $p_1(y\mid x)$ to $p_2(y\mid x)$ defined in \cite{yu2021measuring} is 
\begin{align}
    D_{\varphi}(p_1(y| x)\| p_0(y| x))=D_{\varphi}(C_{xy}^1\| C_{xy}^2)-D_{\varphi}(C_{x}^1\| C_{x}^2),
\end{align}
where $C_{xy}\in \mathcal{S}_+^{p+1}$ is the centered correntropy matrix of the random vector concatenated by $x$ and $y$. $C_{x}\in \mathcal{S}_+^{p}$ is the centered correntropy matrix of the random vector $x$.

Our objective function is composed of four parts, where the first two parts are the factual loss, which measures the fitness of the hypothesis functions to the data of their own group. Minimizing this loss enables the representation and the hypothesis functions we learn to have good predictive capability for the factual outcomes. This loss only uses the factual outcomes, so the data from the treatment group is used to optimize $h_1(\cdot)$ through this loss, and the data from the control group is used to optimize $h_0(\cdot)$ through this loss. Most causal inference methods \cite{shalit2017estimating,shi2019adapting,hassanpour2019counterfactual,hassanpour2019learning,chauhan2023adversarial} that only use this loss will result in the training of $h_1(\cdot)$ and $h_0(\cdot)$ in practice only utilizing data from their respective groups, without considering the applicability of $h_1(\cdot)$ to the control group's data and the applicability of $h_0(\cdot)$ to the treatment group's data. The regularization terms of our objective function take advantage of the predicted counterfactual outcomes, which allows data from different groups to participate in the optimization of $h_1(\cdot)$ and $h_0(\cdot)$ through the regularization term. Minimizing this loss enables the representation we learn to satisfy the unconfoundedness assumption and the hypothesis functions to have good predictive capability for the counterfactual outcomes. Therefore, the representation learned through our objective function is sufficient according to Definition \ref{def1} intuitively.
\begin{algorithm}[tb]
   \caption{CrossNet}
   \label{algorithm1}
\begin{algorithmic}[1]
   \STATE {\bfseries Input:} data $(x_i,y_i,t_i)$, for $i=1,...,n$, the index set $\mathcal{D}_1$, $\mathcal{D}_0$, for treated sample and control sample respectively, tuning parameter $\lambda>0$, loss function $l(\cdot)$, initial representation network $\Phi_I(\cdot)$, initial hypothesis functions $h_{1,I}(\cdot)$, $h_{0,I}(\cdot)$.
   \WHILE {not converged} 
   \FOR{each batch}
   \STATE Calculate the factual loss:\\
   $\mbox{L}_1=\sum_{i=1}l(h_1(\Phi(x_i),y_i),\mbox{ for }i\in \mathcal{D}_1$,\\
   $\mbox{L}_0=\sum_{j=1}l(h_0(\Phi(x_i),y_i),\mbox{ for }j\in \mathcal{D}_0$.
   \STATE Calculate the counterfactual outcomes:\\
   $y_0^1=h_1(\Phi(x_i)),\mbox{ for }i\in \mathcal{D}_0$,\\
   $y_1^0=h_0(\Phi(x_i)),\mbox{ for }i\in \mathcal{D}_1$.
   \STATE Calculate the discrepancy between conditional distributions under treatment and control:\\
   $\mbox{D}_0 = \mbox{Disc}(p(y_0^0| \Phi(x^0)), p(y_1^0| \Phi(x^1)))$,\\
   $\mbox{D}_1 = \mbox{Disc}(p(y_0^1| \Phi(x^0)), p(y_1^1| \Phi(x^1)))$.
   \STATE Update the weights of network through the gradient of the loss $\mbox{L}_1+\mbox{L}_0+\lambda\cdot (\mbox{D}_1+\mbox{D}_0)$.
   \ENDFOR
   \STATE Check the convergence criterion.
   \ENDWHILE
\end{algorithmic}
\end{algorithm}

On the other hand, many representation learning methods \cite{shalit2017estimating,hassanpour2019counterfactual}, impose penalties on the discrepancy between the distributions of $\Phi(\cdot)$ in the treated group and control group, eliminating a certain degree of selection bias by learning a balanced representation. However, the information in $x$ related to $y$ cannot be represented solely by a balanced representation, or in other words, this information is not entirely independent of the treatment variable $t$. Therefore, some studies \cite{hassanpour2019learning,chauhan2023adversarial,curth2021nonparametric} have suggested learning disentangled representations, extracting multiple non-overlapping representations from $x$ to predict $y$ and $t$ separately, along with the penalty on the discrepancy of representation between the treated group and control group. This approach requires learning several non-overlapping representations, which increases the complexity of algorithm optimization. In our framework, we do not impose any direct restrictions on the representation $\Phi(\cdot)$; instead, we start directly from the prediction perspective of two potential outcomes and use different losses to ensure that the representation satisfies sufficiency, thereby enabling the hypothesis functions to have good predictive ability for both factual and counterfactual results, solving a certain problem of selection bias. Now we give the whole procedure for estimating CATE, which is summarized in Algorithm \ref{algorithm1}.

\subsection{Sufficiency}
In the previous subsection, we proposed a new objective function and corresponding algorithm, and intuitively explained the role of each loss in the objective function and its achieved effect. Here, we theoretically demonstrate the rationality of the new method, that is, the representation learned by our method is sufficient.
\begin{theorem}
    Let $\Phi:\mathcal{X}\rightarrow\mathcal{R}$ be a representation function and there exists a sufficient representation in $\mathcal{R}$. Let $h_1(\cdot)$ and $h_0(\cdot)$ be two hypothesis functions that maps $\mathcal{R}$ to $\mathcal{Y}$. If $(\Phi^0(\cdot)$, $h_1^0(\cdot)$, $h_0^0(\cdot))$ is a solution to 
    \begin{align}
    \label{populationobj}
        \min_{\Phi,h_1,h_0}&\mathbb{E}[l(h_1(\Phi(x)),y(1))|t=1]\nonumber\\
        +&\mathbb{E}[l(h_0(\Phi(x)),y(0))|t=0] \nonumber\\
    +&\lambda\cdot D(p(Y(1)| T=1,\Phi(x))\| p(Y(1)|T=0, \Phi(x)))\nonumber \\
    +&\lambda\cdot D(p(Y(0)| T=1,\Phi(x))\| p(Y(0)|T=0,\Phi(x))).
    \end{align}
    Then $\Phi^0(x)$ is a sufficient representation of $x$ for estimating CATE.
\end{theorem}
This theorem indicates that if there exists a sufficient representation of $x$ in the process of estimation for CATE, then the solution obtained by solving the problem \eqref{populationobj} is a sufficient representation. Based on this sufficient representation, subsequent estimation and inference are reliable. The objective function we actually optimize is the empirical form of \eqref{populationobj}.

\section{Experiments}
We apply our approach on three datasets: i) synthetic datasets, where the covariates and outcomes are all simulated so that we know the true CATE; ii) semi-synthetic datasets, Infant Health and Development Program (IHDP), where the covariates are real and the outcomes are simulated, hence we also know the true CATE; iii) real datasets, Jobs, where the covariates and the outcomes are all real, so we do not know the true CATE.
\subsection{Baseline}
We compare our approach with six another approaches based on neural networks: 
\begin{itemize}
    \item \textbf{TNet}: The simplest network to train two hypothesis functions without using the common representation.
    \item \textbf{TARNet, CFRNet}: The two hypothesis functions are trained based on the common (balanced) representation, proposed by \cite{shalit2017estimating}.
    \item \textbf{Dragonnet}: The two hypothesis functions and the propensity score are trained based on the common representation, proposed by \cite{shi2019adapting}.
    \item \textbf{DR-CFR}: Learning three representations for predicting outcomes and treatment variable, proposed by \cite{hassanpour2019learning}.
    \item \textbf{SNet}: Learning five representations for predicting outcomes and treatment variable, proposed by \cite{curth2021nonparametric}.
\end{itemize}
We use the same network architecture in our approach with TARNet and CFRNet, since these three methods all train one representation along with two hypothesis functions. We try to ensure that the model complexity is consistent across different methods to facilitate a fair comparison.

\subsection{Synthetic Dataset}
\begin{figure}[ht]
\begin{center}
\centerline{\includegraphics[width=\columnwidth]{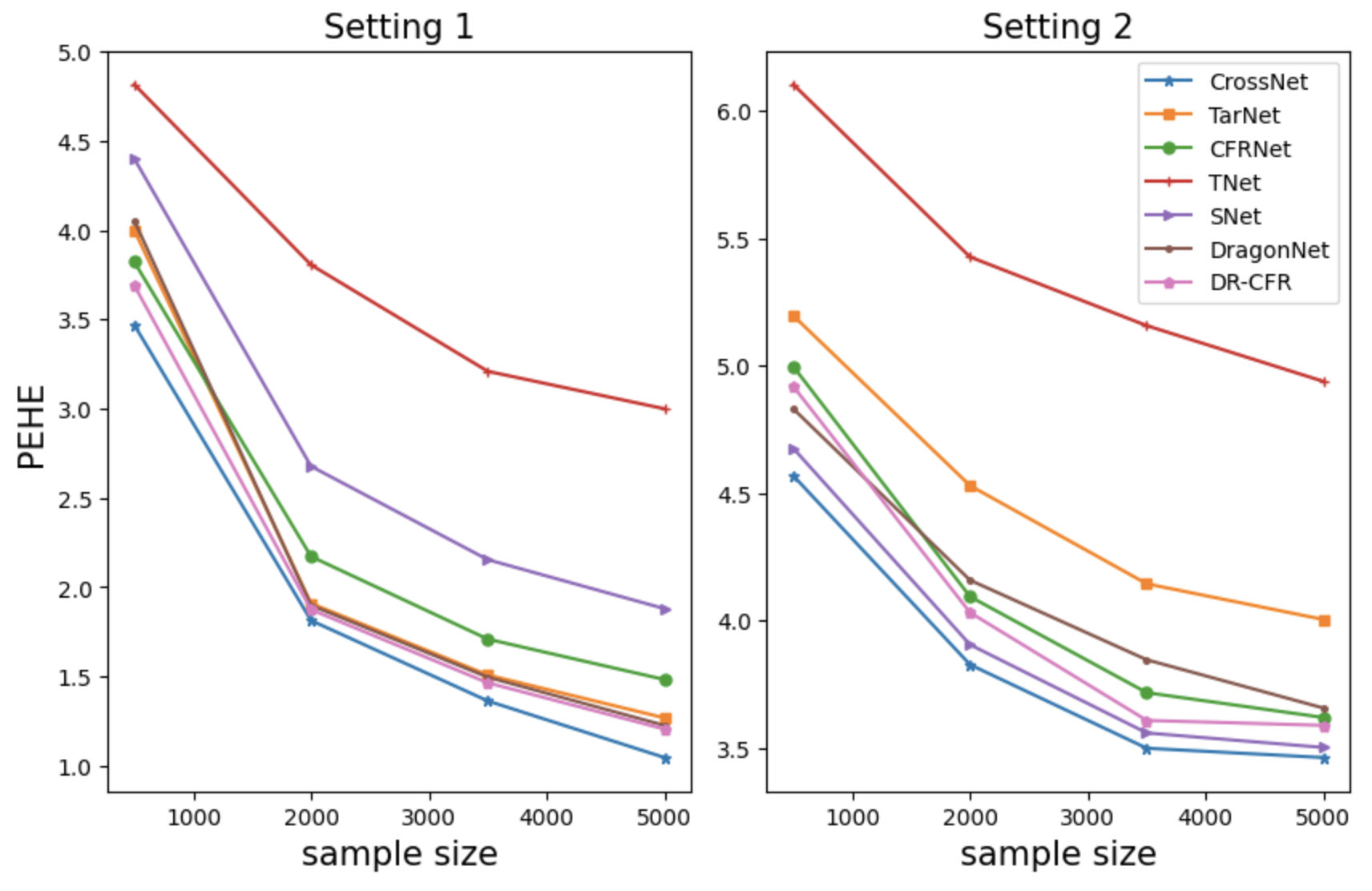}}
\caption{PEHE of all approaches for two simulation settings at different sample size.}
\label{simulation}
\end{center}
\end{figure}

We adopt the same simulation settings as those used in \cite{hassanpour2019learning} and \cite{curth2021nonparametric}. We conduct the simulation on two settings. In both settings, we set the dimension of covariates $x$ to be $d=25$, and $X$ follows a multivariate normal distributions independently among each other, constructed by three disjoint parts $X_c$, $X_o$, $X_t$ with dimension $d_c$, $d_o$ and $d_t$, where $X_c$ are confounders affecting both outcomes and treatment variable, $X_o$ are covariates only affecting the outcomes, $X_t$ are covariates only affecting the treatment variable. Denote $\pi(x)$ as the propensity score. The two simulation settings are:
\begin{itemize}
    \item [i)] $\mu_0(x)=\mu_1(x)=\mathbf{1}^{\top} X_{C O}^2$, where $X_{c o}=\left[X_c, X_o\right]$, $\pi(x)=\operatorname{expit}\left(\xi\left(\frac{1}{d_c} \mathbf{1}^{\top} X_c^2-\omega\right)\right)$, where $\xi$ determines the extent of the selection bias, which is set to be 3. We set $\omega=\operatorname{median}\left(\frac{1}{d_c} \mathbf{1}^{\top} X_c^2\right)$ to center propensity scores. We set $d_c=d_o=5$.
    \item [ii)] We use the same setting as i), except for $\mu_1(x)=\mu_0(x)+\mathbf{1}^{\top}X_\tau ^2$ with another 5 additional covariates $X_\tau$. 
\end{itemize}
For each simulation setting, we conducted experiments with different sample sizes, $n=500,1000,2000,5000$. Each experiment was repeated 10 times to calculate the Precision in the Estimation of Heterogeneous Effects (PEHE) on the test set with size 1000, and the average was taken, where the PEHE is defined as $\mathrm{PEHE}=\sqrt{\frac{1}{N} \sum_{i=1}^N\left(\hat{\mathrm{e}}_i-\mathrm{e}_i\right)^2}$, where $\hat{e}_i = \hat{y}_i^1-\hat{y}_i^0$ and $e_i=y_i^1-y_i^0$. The results are shown in Figure \ref{simulation},
from which, we can see that our approach outperforms other competitive methods in both small sample and large sample scenarios across both two simulation settings.

\subsection{Semi-synthetic Dataset: IHDP}
We conduct experiments on the well known dataset IHDP created by \cite{hill2011bayesian}, where the covariates are real and the outcomes are simulated. The dataset aims to evaluate the effect of home visit from specialist doctors on the cognitive test scores of premature infants. There are 747 units (139 in treated group and 608 in control group) and 25 real covariates, with 6 being continuous and 19 being binary, measuring the features of children and their mothers.  We use the same version as in \cite{curth2021nonparametric}, that is the setting "B" described in \cite{hill2011bayesian}, where the outcomes are simulated by $Y(0) \sim \mathrm{N}\left(\exp \left((X+W) \beta_B\right), 1\right)$ and $Y(1) \sim \mathrm{N}\left(X \beta_B-\omega_B^s, 1\right)$. $\beta_B$ is the coefficient vector with each entry being sampled from $(0, 0.1, 0.2, 0.3, 0.4)$ with probabilities $(0.6, 0.1, 0.1, 0.1, 0.1)$. We use the same 100 replications in \cite{shalit2017estimating,curth2021nonparametric}. Since we know the true treatment effect for all individuals, we use PEHE as the metric to evaluating the performance within-sample and out-of-sample. The results are shown in Table \ref{ihdp}, from which we can see that our proposed approach has the best performance.
\begin{table}[t] 
\begin{center}
\begin{small}
\begin{sc}
\begin{tabular}{lcc}
\toprule
Methods & Within-Sample & Out-of-Sample  \\
\midrule
TNet    & 1.303$\pm$ 0.097& 1.311$\pm$ 0.110 \\
TARNet & 1.294$\pm$ 0.104& 1.299$\pm$ 0.113\\
CFRNet    & 1.156$\pm$ 0.111& 1.162$\pm$ 0.118 \\
DragonNet    & 1.188$\pm$ 0.104& 1.194$\pm$ 0.112 \\
DR-CFR     & 1.334$\pm$ 0.072& 1.342$\pm$ 0.089\\
SNet      & 1.354$\pm$ 0.061& 1.366$\pm$ 0.081 \\
CrossNet      & \textbf{1.077}$\pm$ 0.064& \textbf{1.084}$\pm$ 0.068   \\
\bottomrule
\end{tabular}
\end{sc}
\end{small}
\end{center}
\caption{PEHE of all approaches on the IHDP dataset, averaged across 100 replications along with the standard errors.}
\label{ihdp}
\end{table}

\subsection{Real Dataset: Jobs}
We also apply our approach to the Job dataset first analyzed by \cite{lalonde1986evaluating}. The study aims to evaluating the effect of job training on income and employment status. Following \cite{smith2005does,shalit2017estimating}, we also consider the binary classification task for unemployment. This study contains experimental sample and the PSID comparison group. We split the dataset into train/validation/test sets and repeat 10 times to take average of the results. Since we do not know the true treatment effect, we can not calculate PEHE here. We use another metric defined in \cite{shalit2017estimating}, called policy risk. For a model $f$, the policy risk is $R(f) = 1-(\mathbb{E}[Y_1|\pi_f(x)=1]\cdot p(\pi_f=1)+\mathbb{E}[Y_0|\pi_f(x)=0]\cdot p(\pi_f=0))$, where $\pi_f(x)=1$ if $f(x,1)-f(x,0)>\lambda$, and $\pi_f(x)=0$ otherwise. The policy risk measures the average loss in outcome when using the treatment policy estimated by the model. The results are shown in Table \ref{jobs}, and our approach has the smallest policy risk.
\begin{table}[t]
\begin{center}
\begin{small}
\begin{sc}
\begin{tabular}{lcc}
\toprule
Methods & Within-Sample & Out-of-Sample  \\
\midrule
TNet    & 0.131$\pm$ 0.002& 0.173$\pm$ 0.009 \\
TARNet & 0.119$\pm$ 0.009& 0.139$\pm$ 0.001\\
CFRNet    & 0.111$\pm$ 0.004& 0.140$\pm$ 0.008 \\
DragonNet    & 0.120$\pm$ 0.009& 0.139$\pm$ 0.006 \\
DR-CFR     & 0.117$\pm$ 0.008& 0.138$\pm$ 0.008\\
SNet      & 0.128$\pm$ 0.007& 0.149$\pm$ 0.005 \\
CrossNet      & \textbf{0.109}$\pm$ 0.003& \textbf{0.123}$\pm$ 0.004   \\
\bottomrule
\end{tabular}
\end{sc}
\end{small}
\end{center}
\caption{Policy risk of all approaches on the Jobs dataset, averaged across 10 replications along with the standard errors.}
\label{jobs}
\end{table}

\section{Conclusion}
In this paper we propose a novel neural network approach for estimating CATE. We learn a sufficient representation that satisfies the unconfoundedness assumption by imposing a penalty on the discrepancy between the conditional distributions of two potential outcomes based on the representation under treatment and control, making the final treatment effect estimation more reasonable and effective. Another advantage of this architecture is that it allows the data from different groups to participate in the training of the hypothesis function of the other group, making $h_1$ and $h_0$ more applicable to the data from the other group, thereby mitigating the issue of selection bias. That is, the training of the hypothesis function for the treatment group no longer relies solely on minimizing the factual loss of the treated group; it also takes into account the ability to obtain better counterfactual predictions for the control group's data. This principle applies to the control group as well, thus we fully utilize the information from the samples to learn more robust hypothesis functions. We apply our approach to synthetic dataset, semi-synthetic dataset, real dataset and achieve good performance on all of them.

\section*{Acknowledgments}

This work is supported by Ant Group Research Intern Program. This work is also supported by National key R\&D Programmes of China (2022YFA1003800), National Statistical Science Research Grants of China (2022LD08), the National Natural Science Foundation of China (12231011, 71988101, 71925007 and 72091212), the CAS Project for Young Scientists in Basic Research (YSBR-008).

\bibliographystyle{named}
\bibliography{main}

\end{document}